\pdfoutput=1
\documentclass[11pt]{article}

\usepackage[]{acl}

\usepackage{times}
\usepackage{latexsym}
\usepackage{booktabs}
\usepackage[T1]{fontenc}
\usepackage{bbding}
\usepackage[utf8]{inputenc}
\usepackage{microtype}
\usepackage{inconsolata}

\usepackage{enumitem}
\usepackage{graphicx}
\usepackage{multirow}
\usepackage{arydshln}
\usepackage{listings}
\usepackage{multicol}
\usepackage{enumitem,kantlipsum}
\usepackage{xcolor,colortbl}

\usepackage{hyperref}
\hypersetup{
    colorlinks=true,
    linkcolor=blue,
    filecolor=magenta,      
    urlcolor=blue,
    pdftitle={Overleaf Example},
    pdfpagemode=FullScreen,
    }

\definecolor{my_blue}{HTML}{bae1ff}
\definecolor{my_red}{HTML}{ffb3ba}
\definecolor{my_yellow}{HTML}{ffffba}
\definecolor{my_green}{HTML}{baffc9}
\definecolor{my_orange}{HTML}{FFC99D}
\definecolor{my_purple}{HTML}{e1d4f4}

\title{Can We Rely on LLM Agents to Draft Long-Horizon Plans? \\ \textit{Let's Take TravelPlanner as an Example}}

\author{
    Yanan Chen, Ali Pesaranghader, Tanmana Sadhu, \and Dong Hoon Yi \\
    LG Electronics, Toronto AI Lab, Toronto, Canada \\
    \texttt{\{yanan.chen, ali.pesaranghader, tanmana.sadh, donghoon9.yi\}@lge.com}
}

\begin{document}
\maketitle

% ================================================
%   ABSTRACT
% ================================================

\begin{abstract}

Large language models (LLMs) have brought autonomous agents closer to artificial general intelligence (AGI) due to their promising generalization and emergent capabilities.
There is, however, a lack of studies on how LLM-based agents behave, why they could potentially fail, and how to improve them, particularly in demanding real-world planning tasks.
In this paper, as an effort to fill the gap, we present our study using a realistic benchmark, TravelPlanner \cite{xie2024travelplanner}, where an agent must meet multiple constraints to generate accurate plans. 
We leverage this benchmark to address four key research questions: (1) are LLM agents robust enough to lengthy and noisy contexts when it comes to reasoning and planning? (2) can few-shot prompting adversely impact the performance of LLM agents in scenarios with long context? (3) can we rely on refinement to improve plans, and (4) can fine-tuning LLMs with both positive and \textit{negative} feedback lead to further improvement?
Our comprehensive experiments indicate that, firstly, LLMs often fail to attend to crucial parts of a long context, despite their ability to handle extensive reference information and few-shot examples; secondly, they still struggle with analyzing the long plans and cannot provide accurate feedback for refinement;
thirdly, we propose Feedback-Aware Fine-Tuning (FAFT), which leverages both positive and negative feedback, resulting in substantial gains over Supervised Fine-Tuning (SFT).
Our findings offer in-depth insights to the community on various aspects related to real-world planning applications.

\end{abstract}

% ================================================
%   INTRODUCTION
% ================================================

\section{Introduction}

LLMs have shown significant reasoning and planning results against various benchmarks such as WebArena \cite{zhou2023webarena}, WebShop \cite{yao2022webshop}, AgentBench \cite{liu2023agentbench} and AgentGym \cite{xi2024agentgym} where they act as agents to finish a given task on behalf of humans.
% ------
In this vein, the community considers two main directions for developing LLM-based agents: (1) prompting LLMs for reasoning, planning, and execution \cite{qin2023toolllm, wei2022chain, yao2024tree, wang2022self}, and (2) fine-tuning LLMs for a given task \cite{chen2023fireact, zeng2023agenttuning, zhang2024agentohana, chen2024agent, song2024trial}.
% ------
Despite promising contributions in each direction, it is seen that LLMs still fall short in more complex scenarios. 
TravelPlanner \cite{xie2024travelplanner}, as an example, is a benchmark where an agent should generate a plan which must meet multiple constraints with respect to input queries.
The authors showed that GPT-4-Turbo \cite{openai2023gpt4} could only reach to Final Pass Rate of 4.4\%. This indicates that LLM agents cannot handle long-horizon reasoning and planning.
In this paper, we investigate these challenges further with four research questions using TravelPlanner as the benchmark, and we trust that our promising and negative findings will benefit the community. 

Our extensive experiments indicate that (1) lengthy and noisy context can adversely impact planning ability of the LLM agent, (2) more shots do not necessarily guarantee performance improvement, (3) refinement may not be effective when LLMs are employed as feedback generators; however, it is more likely to work if the feedback generator is based on heuristic rules, and (4) feedback-aware fine-tuning (FAFT), our proposed approach, inspired by negative aware training (NAT) \cite{wang2024learning}, can show remarkable improvement in planning. %[YC-grammar]

\begin{figure}[t]
    \centering
    \resizebox{\linewidth}{!}{
    \includegraphics{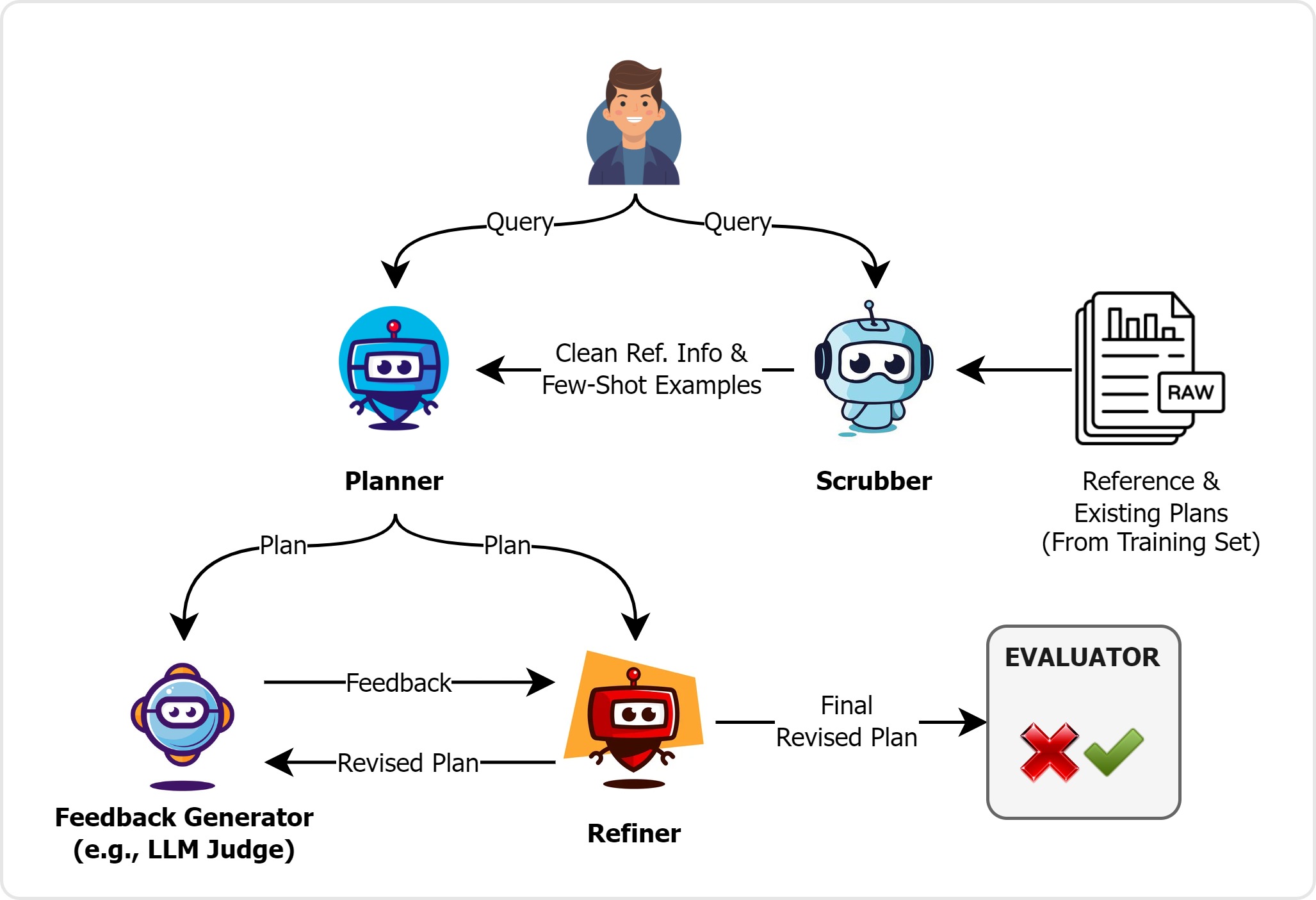}
    }
    \caption{Four LLM agents interact to generate a plan. (Fig.\ \ref{appx:fig:toy-example} is an example for the refinement module.)}
    \label{fig:pipeline}
\end{figure}

% ================================================
%   METHODOLOGY
% ================================================

\section{Methodology}

Our framework, built upon TravelPlanner, consists of five main components: Scrubber, Planner, Feedback Generator, Refiner, and the Evaluation module (as shown in Fig.\ \ref{fig:pipeline}). The scrubber provides clean reference information\footnote{This is a terminology that TravelPlanner uses to refer to necessary information for generating a plan.} and few-shot examples to the Planner for generating a plan. Then, the Feedback Generator provides feedback to the Refiner for improving the plan if required. The interaction continues until the pre-defined settings are met. The Planner is the core of the framework which can be based on either (1) in-context learning (ICL), or (2) supervised fine-tuning (SFT), e.g., FAFT as proposed in Section \ref{findings}-RQ4. Appx.\ \ref{appendix_framework} describes each agent in detail.%[YC-grammar]

\section{Experimental Settings}

\noindent \textbf{Basic Setting.}
Since the focus of our work is on agents' capabilities in drafting plans, we only rely on the \textit{Sole Planning} setting from TravelPlanner. That is, all comprehensive and necessary information, which are human annotations, is directly provided to the planner agent.
We also consider the \textbf{Direct}\footnote{the query is input directly into the model along with instructions detailing the task and relevant information gathered.} planning strategy for its simplicity because it performs at a similar level to other reasoning techniques such as ZS-CoT \cite{wei2022chain}, ReAct \cite{yao2022react} and Reflexion \cite{shinn2024reflexion}.

\vspace{0.5em}
\noindent \textbf{Dataset.} (See Appx.\ \ref{appendix_dataset})
We use the training set for both few-shot prompting and fine-tuning because it provides annotated plans.
We evaluate the agent against both \textit{validation} and \textit{test sets} for RQ1 and RQ2 in Section \ref{findings}. As for RQ3, we consider only the validation set because we do not have access to the system feedback offline.
Regarding RQ4, we use the training set for fine-tuning the (Open-LLM) planner agent.

\vspace{0.5em}
\noindent \textbf{Metrics.} 
We utilize the original evaluation metrics from TravelPlanner, which evaluate performance based on the pass rates of multiple constraints. Additional details are available in Appx.\ \ref{appendix_eval}.
% ================================================
%   FINDINGS
% ================================================

\begin{table*}[ht]
    \setlength\tabcolsep{3pt}
    \resizebox{\linewidth}{!}{
            \begin{tabular}{cccccccccccccccc}
            \toprule
            \multicolumn{16}{c}{\textbf{GPT-3.5-Turbo as Planner}} \\ \hline
            &  & \multicolumn{7}{c}{\textbf{Validation Set} (\#180)} & \multicolumn{7}{c}{\textbf{Test Set} (\#1,000)} \\ \cmidrule(l){3-9} \cmidrule(l){10-16}
            \multirow{2}{*}{\begin{tabular}[c]{@{}c@{}}Reference \\ Scrubbed?\end{tabular}} & \multirow{2}{*}{\begin{tabular}[c]{@{}c@{}}Num. \\ Shots\end{tabular}} & \multirow{2}{*}{\begin{tabular}[c]{@{}c@{}}Delivery \\ Rate\end{tabular}} & \multicolumn{2}{c}{\begin{tabular}[c]{@{}c@{}}Commonsense \\ Pass Rate\end{tabular}} & \multicolumn{2}{c}{\begin{tabular}[c]{@{}c@{}}Hard Constraint\\ Pass Rate\end{tabular}} & \multirow{2}{*}{\begin{tabular}[c]{@{}c@{}}Final \\ Pass Rate\end{tabular}} & \multirow{2}{*}{\begin{tabular}[c]{@{}c@{}}Halluc. \\ Rate\end{tabular}} & \multirow{2}{*}{\begin{tabular}[c]{@{}c@{}}Delivery \\ Rate\end{tabular}} & \multicolumn{2}{c}{\begin{tabular}[c]{@{}c@{}}Commonsense \\ Pass Rate\end{tabular}} & \multicolumn{2}{c}{\begin{tabular}[c]{@{}c@{}}Hard Constraint\\ Pass Rate\end{tabular}} & \multirow{2}{*}{\begin{tabular}[c]{@{}c@{}}Final \\ Pass Rate\end{tabular}} & \multirow{2}{*}{\begin{tabular}[c]{@{}c@{}}Halluc. \\ Rate\end{tabular}} \\ \cmidrule(l){4-5} \cmidrule(l){6-7} \cmidrule(l){11-12} \cmidrule(l){13-14} 
            % ----------------------------------------------------------
            (RQ1) & (RQ2) & & Micro & Macro & Micro & Macro & & & & Micro & Macro & Micro & Macro & & \\ \hline
            No & 0 & 100 & 60.2 & 4.4 & 11.0 & 2.8 & 0.0 & 57.4 & 100 & 60.8 & 3.5 & 13.6 & 4.9 & 0.6 & 61.1 \\
            No & 1 & 100 & 65.4 & 11.0 & 17.5 & 5.1 & 1.0 & 52.3 & 100 & 64.0 & 10.1 & 16.1 & 6.4 & 1.2 & 59.4 \\ % \hdashline
            Yes & 0 & 100 & 74.4 & 18.9 & 29.0 & 14.4 & 4.4 & 41.6 & 100 & 70.3 & 12.3 & 25.0 & 10.7 & 2.7 & 49.8 \\
            Yes & 1 & 100 & 80.6 & 24.4 & 40.2 & 17.8 & 7.2 & \textbf{35.5} & 100 & 78.0 & 18.6 & 36.1 & 17.7 & 4.9 & \textbf{40.8} \\
            Yes & 2 & \textbf{100} & \textbf{82.6} & \textbf{32.2} & \textbf{41.2} & \textbf{17.8} & \textbf{7.2} & 38.8 & \textbf{100} & \textbf{80.9} & \textbf{22.4} & \textbf{34.3} & \textbf{16.7} & \textbf{6.5} & 44.3 \\
            Yes & 4 & 100 & 81.5 & 29.4 & 35.5 & 12.2 & 5.8 & 46.6 & 100 & 80.3 & 21.1 & 31.5 & 15.0 & 5.2 & 50.8 \\
            Yes & 5 & 100 & 81.1 & 26.3 & 32.6 & 12.4 & 4.8 & 49.8 & 100 & 79.5 & 20.4 & 30.4 & 13.2 & 5.6 & 53.1 \\ \bottomrule
        \end{tabular}
    }
    \caption{Performance of GPT-3.5-Turbo as the Planner agent for different settings for RQ1 and RQ2}
    \label{tab:rq12}
\end{table*}

\begin{table*}[ht]
    \centering
    \resizebox{\linewidth}{!}{
        \begin{tabular}{lcccccccccc}
        \cmidrule(l){3-11}
        & & \multicolumn{9}{c}{\textbf{GPT-3.5-Turbo as Planner and GPT-4-Turbo as Refiner}} \\
        & & \multicolumn{9}{c}{\textbf{vs. Validation Set (\#180)}} \\ \hline
        \multirow{2}{*}{\begin{tabular}[l]{@{}l@{}} Feedback Generator \\ (RQ3)\end{tabular}} & \multirow{2}{*}{\begin{tabular}[c]{@{}c@{}}Refinement \\ Iteration\end{tabular}} & \multirow{2}{*}{\begin{tabular}[c]{@{}c@{}}Delivery \\ Rate\end{tabular}} & \multicolumn{2}{c}{\begin{tabular}[c]{@{}c@{}}Commonsense\\ Pass Rate\end{tabular}} & \multicolumn{2}{c}{\begin{tabular}[c]{@{}c@{}}Hard Constraint\\ Pass Rate\end{tabular}} & \multirow{2}{*}{\begin{tabular}[c]{@{}c@{}}Final\\ Pass Rate\end{tabular}} & \multirow{2}{*}{\begin{tabular}[c]{@{}c@{}}Uplift \\ Ratio ($\uparrow$)\end{tabular}} & \multirow{2}{*}{\begin{tabular}[c]{@{}c@{}}Flat\\ Ratio ($\downarrow$)\end{tabular}} & \multirow{2}{*}{\begin{tabular}[c]{@{}c@{}}Downgrade\\ Ratio ($\downarrow$) \end{tabular}} \\ \cmidrule(l){4-5} \cmidrule(l){6-7} 
        & & & \multicolumn{1}{c}{Micro} & \multicolumn{1}{c}{Macro} & \multicolumn{1}{c}{Micro} & \multicolumn{1}{c}{Macro} & & & & \\ \toprule
        None & 0 & 100 & 82.6 & 32.2 & 41.2 & 17.8 & \cellcolor{my_red}{7.2} & \multicolumn{1}{c}{--} & \multicolumn{1}{c}{--} & \multicolumn{1}{c}{--} \\ \midrule
        \multirow{4}{*}{Oracle (Heuristic Rules)} & 1 & 100 & 89.7 & 51.1 & 50.0 & 22.2 & 11.7 & 46.1 & 52.8 & 1.1 \\
         & 2 & 100 & 89.0 & 54.4 & 50.5 & 18.3 & 12.8 & 8.9 & 76.7 & 14.4 \\
         & 3 & 100 & 89.9 & 56.1 & 50.7 & 21.1 & 13.3 & 13.9 & 78.9 & 7.2 \\
         & 4 & 100 & 89.1 & 59.4 & 49.8 & 21.7 & \cellcolor{my_blue}{\textbf{13.9}} & 5.0 & 86.7 & 8.3 \\ \midrule
        \multirow{4}{*}{Random} & 1 & 100 & 82.3 & 31.1 & 47.1 & 21.1 & 7.2 & 21.7 & 53.3 & 25.0 \\
         & 2 & 100 & 82.3 & 32.2 & 46.2 & 20.0 & \cellcolor{yellow}{8.3} & 18.3 & 63.9 & 17.8 \\
         & 3 & 100 & 82.0 & 30.6 & 45.5 & 20.0 & 7.2 & 19.4 & 61.7 & 18.9 \\
         & 4 & 100 & 82.6 & 30.6 & 44.8 & 18.3 & 7.2 & 18.9 & 62.8 & 18.3 \\ \midrule
         \multirow{4}{*}{GPT-3.5-Turbo (0125)} & 1 & 100 & 82.0 & 24.4 & 41.4 & 21.7 & \cellcolor{yellow}{8.9} & 22.2 & 52.8 & 25.0 \\
         & 2 & 100 & 82.9 & 28.9 & 40.9 & 18.3 & 8.9 & 24.4 & 55.0 & 20.6 \\
         & 3 & 100 & 83.8 & 27.8 & 41.7 & 20.0 & 8.9 & 25.0 & 56.1 & 18.9 \\
         & 4 & 100 & 82.4 & 26.7 & 40.7 & 18.9 & 7.8 & 19.4 & 57.8 & 22.8 \\ \midrule
        \multirow{4}{*}{GPT-4-Turbo (1106-preview)} & 1 & 100 & 86.9 & 32.8 & 39.3 & 20.0 & \cellcolor{yellow}{9.4} & 34.4 & 40.6 & 25.0 \\
         & 2 & 100 & 84.3 & 29.4 & 37.9 & 15.6 & 7.2 & 20.0 & 59.4 & 20.6 \\
         & 3 & 100 & 84.6 & 30.0 & 40.5 & 18.3 & 7.2 & 18.3 & 67.2 & 14.4 \\
         & 4 & 100 & 86.4 & 28.3 & 37.9 & 20.0 & 6.7 & 19.4 & 58.3 & 22.2 \\
        \bottomrule
        \end{tabular}
        }
    \caption{Performance of different Feedback Generators. Uplift Ratio ($\uparrow$), Flat Ratio ($\downarrow$), and Downgrade Ratio ($\downarrow$) show what percentage of plans has improved, not changed, and deteriorated, respectively.}
    \label{tab:rq3}
\end{table*}

\section{Findings}
\label{findings}

% ====================> RQ1 <==========================

\textbf{RQ1: Are LLM agents robust enough to noisy information for reasoning and planning?}
\noindent Table \ref{tab:rq12} shows that GPT-3.5-Turbo has a better performance when it receives \emph{shrunk} reference information. This indicates that GPT-3.5-Turbo still struggles to attend to the most important parts of a given context for reasoning, prone to excessive irrelevant (context) chunks. Therefore, it is worth considering an external intelligent context-cleaning agent. 

% ====================> RQ2 <==========================

\vspace{0.5em}
\noindent \textbf{RQ2}: \textbf{Can more shots help with the planning task, or does it worsen hallucination?}
It is commonly accepted that having more few-shots is helpful in ICL, but does it apply to TravelPlanner?
As Table \ref{tab:rq12} shows, the Final Pass rate reaches its highest value when there are $2$ shots, while having more shots may not improve if not hurt more.
We presume that more shots in the context window may distract the LLM and lead to hallucination (e.g., using entities that do not exist in the given reference information).
The results of the Hallucination Rate attest to this assumption.
That is, giving more shots may potentially cause severer hallucination in tasks where the context of the reference information is complex tabular texts. Another finding is that at least one in-context example is beneficial \cite{xie2022does}. 
Finally, we conclude that as we have more shots, the pass rate and hallucination rate results worsen.

\vspace{0.5em}
\noindent 
Our RQ1 and RQ2 observations align with the existing theoretical and experimental works, e.g., \cite{han2023lm, levy2024same}, which identify the potential causes underlying current LLMs' failure in length generalization, that when they encounter a much longer context, the attention scores are diluted, and thus the score distribution becomes flat leading to information loss. 
That is, the entropy of the attention score will explode with increasing context.
In other words, LLMs become lost in how to focus on the right information, especially when pre-training is done on shorter text segments.
% ====================> RQ3 <==========================

\vspace{0.5em}
\noindent \textbf{RQ3: Can we rely on refinement to improve plans?}
To address this, we require feedback that highlights what went wrong, accompanied by explanations of the reasons behind the issues. For that, we examine the reliability of GPT-3.5-Turbo and GPT-4-Turbo as LLM-based feedback generators. In addition, as an ablative point of view, we also consider \textit{Random} and \textit{Oracle} feedback generators. The former refers to the setting where we fabricate the feedback using random content in a valid format, and the latter uses heuristic hard-coded rules\footnote{Rules from TravelPlanner: \url{https://github.com/OSU-NLP-Group/TravelPlanner/tree/main/evaluation}} to check whether the plan complies with the constraints.
Furthermore, we do not consider any weaker language models because they have shown to be incapable of handling this kind of task in previous studies \cite{madaan2024self}. 
We only focus on commonsense constraints in this part due to the frequent absence of hard constraints in the queries and the feedback\footnote{
Consistent with the original setting of TravelPlanner, i.e., plans that fail to satisfy all commonsense constraints will not proceed to receive feedback regarding hard constraints. This decision is rooted in the dependency of hard constraint computation on commonsense criteria.}.
We present the results for RQ3 in Table \ref{tab:rq3}.
The feedback generator and the Refiner agent interact iteratively; at each iteration, the previously generated plans are reviewed by the feedback generator to draft feedback subjectively. Considering the feedback, if any refinement is needed, i.e., any constraint is not met, the Refiner agent is triggered to modify the plan.
\emph{This design simulates the production-level environment where no Oracle feedback generator is available to check whether a plan needs refinement.} %[YC-grammar]
We summarize our findings as follows:

\begin{itemize}[noitemsep, topsep=0em, leftmargin=*]

\item \textit{Refinement can help improve the plans if the feedback is of high quality and precise} -- 
We see that the refinement helps with improving the plans if the feedback is accurate and well-organized as in the Oracle setting. 
The table shows, in the first iteration, $46.1\%$ of the plans are improved and the final pass rate is lifted from $7.2\%$ to $11.7\%$. From the second iteration, we do not see any significant improvement and the pass rates saturate.  %[YC-grammar]

\item \textit{LLM feedback generators are not reliable} --
The LLM-based feedback generators, equipped with meticulously designed prompts with two-shots, still struggle with writing unerring feedback.
Specifically, for faulty plans, these feedback generators cannot identify where the violation is or write excessive (baseless) feedback.
Additionally, for qualified plans, they may generate false negative feedback which triggers the refinement module and causes unnecessary modification potentially leading to an invalid plan. As a result, the overall performance becomes stagnant, i.e., the Flat Ratio dominates, and modifications in a negative direction (downgrade ratio) have counteracted the positive changes (uplift ratio). %[YC-grammar]
\end{itemize} 

% ====================> RQ4 <==========================

\vspace{0.5em}
\noindent \textbf{RQ4: Can we enhance the development of a superior planner by employing our feedback-aware fine-tuning (FAFT) technique, as opposed to relying on off-the-shelf proprietary LLMs?} %[YC]
For plan generation, we can use the Oracle feedback for in-context learning (as shown in RQ3) or fine-tuning an (open-source) LLM. 
The focus of this experiment lies in the latter aspect, where we examine the performance of SFT and FAFT in building the Planner agent.

\vspace{0.5em}
\noindent \textbf{SFT vs.\ FAFT} --
In our proposed approach, i.e., FAFT, we extend beyond the considerations of query, reference information, and annotated plan as in SFT, by also incorporating feedback into the fine-tuning process (Appx.\ \ref{appendix_sft_faft}).
To generate feedback, we initially use the queries from the training set and prompt the Planner agent to generate plans\footnote{Under the same setting as in RQ1 and RQ2.}. 
Subsequently, we gather feedback by evaluating the generated plans using the Oracle (i.e., the system).
We set \verb|temperature| to $1.0$ for plan generation to enhance the diversity of the generated plans, consequently introducing a broader range of positive and negative feedback samples.
We collected $14,800$ samples including $45$ original annotated plans from the training set together with their  \verb|all-success| feedback\footnote{It is noteworthy that more samples could be collected.} (Appx.\ \ref{feedback_example_oracle}). 
During inference, in the prompt, the feedback will be set to \verb|all-success|, aiming to encourage the model to generate a correct plan.
Appx.\ \ref{nat_sample_inference} provides more information.

Table \ref{tab:rq4} presents the impact of FAFT where a significant improvement is witnessed across all pass rates, compared to Vanilla Llama-3-8B and its SFT version.
This observation aligns with the previous studies, e.g., Negative-Aware Training (NAT) \cite{wang2024learning}, that the performance can be boosted by increasing the diversity of prompts. In FAFT, elaborative and rich feedback acts as thought chains to improve the agent's planning.
Further, our results validate the recent works \cite{lee2023teaching,wei2023magicoder} by suggesting that (1) injecting auxiliary information in conventional SFT data can markedly improve the performance, 
and 
(2) a CoT-style training set and detailed scratchpads can significantly improve learning by reducing sample complexity.

Our findings advocate that when annotation is scarce while interaction with the system is affordable, collecting samples with comprehensive and rich feedback (either positive or negative), can be worthwhile. This approach can be seen as a promising alternative to RL-based solutions, such as PPO \cite{schulman2017proximal}, which has been criticized for instability.
 
\begin{table}[t]
    \resizebox{\columnwidth}{!}{
    \centering
    \begin{tabular}{lcccccc}
        \toprule
     \multicolumn{7}{c}{\textbf{Llama-3-8B as Planner}} \\ \hline
       \multirow{2}{*}{\begin{tabular}[c]{@{}c@{}}Planner \\ (RQ4) \end{tabular}} & \multirow{2}{*}{\begin{tabular}[c]{@{}c@{}}Delivery \\ Rate\end{tabular}} & \multicolumn{2}{c}{\begin{tabular}[c]{@{}c@{}}Commonsense \\ Pass Rate\end{tabular}} & \multicolumn{2}{c}{\begin{tabular}[c]{@{}c@{}}Hard Constraint\\ Pass Rate\end{tabular}} & \multirow{2}{*}{\begin{tabular}[c]{@{}c@{}}Final \\ Pass Rate\end{tabular}} \\ \cmidrule(l){3-4} \cmidrule(l){5-6}
        & & Micro & Macro & Micro & Macro & \\ \hline
        Vanilla & 94.4 & 49.5 & 1.1 & 7.9 & 0.0 & 0.0 \\ 
        + SFT & 97.8 & 64.2 & 11.1 & 12.4 & 6.1 & 3.9 \\
        + FAFT & \textbf{98.9} & \textbf{81.7} & \textbf{28.9} & \textbf{36.9} &  \textbf{15.0} & \textbf{8.3} \\
        \bottomrule
    \end{tabular}
    }
    \caption{Performance of Llama-3-8B +SFT and +FAFT.
    % test set. 
    % \footnote{We will release the training set held on huggingface hub upon acceptance.} 
    }
    \label{tab:rq4}
\end{table}

\section{Conclusion}

In this paper, we studied the impacts of context, the number of shots, and the utilization of feedback on a complex long-horizon planning task known as TravelPlanner. Our findings aim to advance a broader spectrum of agentic frameworks and strategies within the research community.

For future work, we plan to explore methods that incorporate annotated shots in SFT and post-training. This approach can address the bottleneck where LLMs' knowledge and skills are predominantly acquired during pre-training, while alignment SFT teaches the model which sub-distribution of formats to use when interacting with users \cite{zhou2024lima}. Finally, we will explore the interplay between RLHF and FAFT.

% \clearpage

\section*{Limitations}

Due to budget constraints, we were only able to use GPT-3.5-Turbo as the Planner agent for RQ1 and RQ2. For RQ4, further investigations are needed to explore the relationship between the magnitude of gains and the size of the FAFT training set, as well as the impact of the ratio of positive to negative samples on the final performance. Additionally, enhancing the feedback expressions could further improve the performance of FAFT. It would also be interesting to investigate RLHF techniques, such as DPO \cite{rafailov2024direct} and PRO \cite{song2024preference}, to better utilize feedback.

\section*{Ethics Statement}

Our work is founded upon TravelPlanner, a benchmark designed for complex planning tasks. We adhere to the original work's specifications, utilizing their data, evaluation scripts, and definitions of commonsense. Acknowledging the foundational concepts and designs of the original benchmark, we strictly adhere to TravelPlanner's guidelines, ensuring the integrity of the evaluation process by prohibiting any form of cheating in the validation and test sets. This commitment upholds the fairness and reliability of this work.

\vspace{0.5em}
\noindent As for environmental cost, we acknowledge that our work necessitated extensive experiments to derive robust conclusions. However, future endeavours can leverage these insights, potentially reducing the need for numerous large-scale comparisons. Models intended for production could undergo training once, utilizing the most promising settings identified through our research.

% \section*{Acknowledgements}

\bibliographystyle{acl_natbib}
\bibliography{arxiv}

\begin{thebibliography}{60}
\expandafter\ifx\csname natexlab\endcsname\relax\def\natexlab#1{#1}\fi

\bibitem[{Chang et~al.(2024)Chang, Wang, Wang, Wu, Yang, Zhu, Chen, Yi, Wang, Wang et~al.}]{chang2024survey}
Yupeng Chang, Xu~Wang, Jindong Wang, Yuan Wu, Linyi Yang, Kaijie Zhu, Hao Chen, Xiaoyuan Yi, Cunxiang Wang, Yidong Wang, et~al. 2024.
\newblock A survey on evaluation of large language models.
\newblock \emph{ACM Transactions on Intelligent Systems and Technology}, 15(3):1--45.

\bibitem[{Chen et~al.(2023{\natexlab{a}})Chen, Scheurer, Korbak, Campos, Chan, Bowman, Cho, and Perez}]{chen2023improving}
Angelica Chen, J{\'e}r{\'e}my Scheurer, Tomasz Korbak, Jon~Ander Campos, Jun~Shern Chan, Samuel~R Bowman, Kyunghyun Cho, and Ethan Perez. 2023{\natexlab{a}}.
\newblock Improving code generation by training with natural language feedback.
\newblock \emph{arXiv preprint arXiv:2303.16749}.

\bibitem[{Chen et~al.(2023{\natexlab{b}})Chen, Shu, Shareghi, Collier, Narasimhan, and Yao}]{chen2023fireact}
Baian Chen, Chang Shu, Ehsan Shareghi, Nigel Collier, Karthik Narasimhan, and Shunyu Yao. 2023{\natexlab{b}}.
\newblock Fireact: Toward language agent fine-tuning.
\newblock \emph{arXiv preprint arXiv:2310.05915}.

\bibitem[{Chen et~al.(2024)Chen, Liu, Wang, Zhang, Liu, Lin, Chen, and Zhao}]{chen2024agent}
Zehui Chen, Kuikun Liu, Qiuchen Wang, Wenwei Zhang, Jiangning Liu, Dahua Lin, Kai Chen, and Feng Zhao. 2024.
\newblock Agent-flan: Designing data and methods of effective agent tuning for large language models.
\newblock \emph{arXiv preprint arXiv:2403.12881}.

\bibitem[{Christianos et~al.(2023)Christianos, Papoudakis, Zimmer, Coste, Wu, Chen, Khandelwal, Doran, Feng, Liu et~al.}]{christianos2023pangu}
Filippos Christianos, Georgios Papoudakis, Matthieu Zimmer, Thomas Coste, Zhihao Wu, Jingxuan Chen, Khyati Khandelwal, James Doran, Xidong Feng, Jiacheng Liu, et~al. 2023.
\newblock Pangu-agent: A fine-tunable generalist agent with structured reasoning.
\newblock \emph{arXiv preprint arXiv:2312.14878}.

\bibitem[{Deng et~al.(2023)Deng, Gu, Zheng, Chen, Stevens, Wang, Sun, and Su}]{deng2023mind2web}
Xiang Deng, Yu~Gu, Boyuan Zheng, Shijie Chen, Samuel Stevens, Boshi Wang, Huan Sun, and Yu~Su. 2023.
\newblock \href {http://arxiv.org/abs/2306.06070} {Mind2web: Towards a generalist agent for the web}.

\bibitem[{Fei et~al.(2023)Fei, Niu, Zhou, Hou, Bai, Deng, and Han}]{fei2023extending}
Weizhi Fei, Xueyan Niu, Pingyi Zhou, Lu~Hou, Bo~Bai, Lei Deng, and Wei Han. 2023.
\newblock Extending context window of large language models via semantic compression.
\newblock \emph{arXiv preprint arXiv:2312.09571}.

\bibitem[{gkamradt(2023)}]{LLMTest_NeedleInAHaystack}
gkamradt. 2023.
\newblock Llmtest needle in a haystack - pressure testing llms.
\newblock \url{https://github.com/gkamradt/LLMTest_NeedleInAHaystack}.

\bibitem[{Guo et~al.(2024)Guo, Chen, Wang, Chang, Pei, Chawla, Wiest, and Zhang}]{guo2024large}
Taicheng Guo, Xiuying Chen, Yaqi Wang, Ruidi Chang, Shichao Pei, Nitesh~V Chawla, Olaf Wiest, and Xiangliang Zhang. 2024.
\newblock Large language model based multi-agents: A survey of progress and challenges.
\newblock \emph{arXiv preprint arXiv:2402.01680}.

\bibitem[{Han et~al.(2023)Han, Wang, Xiong, Chen, Ji, and Wang}]{han2023lm}
Chi Han, Qifan Wang, Wenhan Xiong, Yu~Chen, Heng Ji, and Sinong Wang. 2023.
\newblock Lm-infinite: Simple on-the-fly length generalization for large language models.
\newblock \emph{arXiv preprint arXiv:2308.16137}.

\bibitem[{Kim et~al.(2024)Kim, Baldi, and McAleer}]{kim2024language}
Geunwoo Kim, Pierre Baldi, and Stephen McAleer. 2024.
\newblock Language models can solve computer tasks.
\newblock \emph{Advances in Neural Information Processing Systems}, 36.

\bibitem[{Lee et~al.(2023)Lee, Sreenivasan, Lee, Lee, and Papailiopoulos}]{lee2023teaching}
Nayoung Lee, Kartik Sreenivasan, Jason~D Lee, Kangwook Lee, and Dimitris Papailiopoulos. 2023.
\newblock Teaching arithmetic to small transformers.
\newblock \emph{arXiv preprint arXiv:2307.03381}.

\bibitem[{Levy et~al.(2024)Levy, Jacoby, and Goldberg}]{levy2024same}
Mosh Levy, Alon Jacoby, and Yoav Goldberg. 2024.
\newblock Same task, more tokens: the impact of input length on the reasoning performance of large language models.
\newblock \emph{arXiv preprint arXiv:2402.14848}.

\bibitem[{Li et~al.(2024)Li, Hammoud, Itani, Khizbullin, and Ghanem}]{li2024camel}
Guohao Li, Hasan Hammoud, Hani Itani, Dmitrii Khizbullin, and Bernard Ghanem. 2024.
\newblock Camel: Communicative agents for" mind" exploration of large language model society.
\newblock \emph{Advances in Neural Information Processing Systems}, 36.

\bibitem[{Li et~al.(2023)Li, Yuan, Feng, Pan, Sun, Wang, Wang, and Li}]{Li2023TurningDI}
Yiwei Li, Peiwen Yuan, Shaoxiong Feng, Boyuan Pan, Bin Sun, Xinglin Wang, Heda Wang, and Kan Li. 2023.
\newblock \href {https://api.semanticscholar.org/CorpusID:266375154} {Turning dust into gold: Distilling complex reasoning capabilities from llms by leveraging negative data}.
\newblock In \emph{AAAI Conference on Artificial Intelligence}.

\bibitem[{Liu et~al.(2023{\natexlab{a}})Liu, Li, Xiang, Wang, and Qian}]{liu2023tcra}
Junyi Liu, Liangzhi Li, Tong Xiang, Bowen Wang, and Yiming Qian. 2023{\natexlab{a}}.
\newblock Tcra-llm: Token compression retrieval augmented large language model for inference cost reduction.
\newblock In \emph{Findings of the Association for Computational Linguistics: EMNLP 2023}, pages 9796--9810.

\bibitem[{Liu et~al.(2023{\natexlab{b}})Liu, Yu, Zhang, Xu, Lei, Lai, Gu, Ding, Men, Yang, Zhang, Deng, Zeng, Du, Zhang, Shen, Zhang, Su, Sun, Huang, Dong, and Tang}]{liu2023agentbench}
Xiao Liu, Hao Yu, Hanchen Zhang, Yifan Xu, Xuanyu Lei, Hanyu Lai, Yu~Gu, Hangliang Ding, Kaiwen Men, Kejuan Yang, Shudan Zhang, Xiang Deng, Aohan Zeng, Zhengxiao Du, Chenhui Zhang, Sheng Shen, Tianjun Zhang, Yu~Su, Huan Sun, Minlie Huang, Yuxiao Dong, and Jie Tang. 2023{\natexlab{b}}.
\newblock Agentbench: Evaluating llms as agents.
\newblock \emph{arXiv preprint arXiv: 2308.03688}.

\bibitem[{Liu et~al.(2024)Liu, Yao, Zhang, Yang, Liu, Tan, Choubey, Lan, Wu, Wang et~al.}]{liu2024agentlite}
Zhiwei Liu, Weiran Yao, Jianguo Zhang, Liangwei Yang, Zuxin Liu, Juntao Tan, Prafulla~K Choubey, Tian Lan, Jason Wu, Huan Wang, et~al. 2024.
\newblock Agentlite: A lightweight library for building and advancing task-oriented llm agent system.
\newblock \emph{arXiv preprint arXiv:2402.15538}.

\bibitem[{Ma et~al.(2024)Ma, Zhang, Zhu, Yang, Yang, Jin, Lan, Kong, and He}]{ma2024agentboard}
Chang Ma, Junlei Zhang, Zhihao Zhu, Cheng Yang, Yujiu Yang, Yaohui Jin, Zhenzhong Lan, Lingpeng Kong, and Junxian He. 2024.
\newblock Agentboard: An analytical evaluation board of multi-turn llm agents.
\newblock \emph{arXiv preprint arXiv:2401.13178}.

\bibitem[{Madaan et~al.(2024)Madaan, Tandon, Gupta, Hallinan, Gao, Wiegreffe, Alon, Dziri, Prabhumoye, Yang et~al.}]{madaan2024self}
Aman Madaan, Niket Tandon, Prakhar Gupta, Skyler Hallinan, Luyu Gao, Sarah Wiegreffe, Uri Alon, Nouha Dziri, Shrimai Prabhumoye, Yiming Yang, et~al. 2024.
\newblock Self-refine: Iterative refinement with self-feedback.
\newblock \emph{Advances in Neural Information Processing Systems}, 36.

\bibitem[{OpenAI(2023)}]{openai2023gpt4}
OpenAI. 2023.
\newblock \href {https://arxiv.org/abs/2303.08774} {Gpt-4 technical report}.
\newblock \emph{arXiv preprint arXiv:2303.08774}.

\bibitem[{Pan et~al.(2024)Pan, Zhang, Tomlin, Zhou, Levine, and Suhr}]{pan2024autonomous}
Jiayi Pan, Yichi Zhang, Nicholas Tomlin, Yifei Zhou, Sergey Levine, and Alane Suhr. 2024.
\newblock Autonomous evaluation and refinement of digital agents.
\newblock \emph{arXiv preprint arXiv:2404.06474}.

\bibitem[{Paul et~al.(2023)Paul, Ismayilzada, Peyrard, Borges, Bosselut, West, and Faltings}]{paul2023refiner}
Debjit Paul, Mete Ismayilzada, Maxime Peyrard, Beatriz Borges, Antoine Bosselut, Robert West, and Boi Faltings. 2023.
\newblock Refiner: Reasoning feedback on intermediate representations.
\newblock \emph{arXiv preprint arXiv:2304.01904}.

\bibitem[{Qian et~al.(2024)Qian, Liu, Zhang, Mao, Zhou, Chen, and Dou}]{qian2024long}
Hongjin Qian, Zheng Liu, Peitian Zhang, Kelong Mao, Yujia Zhou, Xu~Chen, and Zhicheng Dou. 2024.
\newblock Are long-llms a necessity for long-context tasks?
\newblock \emph{arXiv preprint arXiv:2405.15318}.

\bibitem[{Qin et~al.(2023)Qin, Liang, Ye, Zhu, Yan, Lu, Lin, Cong, Tang, Qian et~al.}]{qin2023toolllm}
Yujia Qin, Shihao Liang, Yining Ye, Kunlun Zhu, Lan Yan, Yaxi Lu, Yankai Lin, Xin Cong, Xiangru Tang, Bill Qian, et~al. 2023.
\newblock Toolllm: Facilitating large language models to master 16000+ real-world apis.
\newblock \emph{arXiv preprint arXiv:2307.16789}.

\bibitem[{Rafailov et~al.(2024)Rafailov, Sharma, Mitchell, Manning, Ermon, and Finn}]{rafailov2024direct}
Rafael Rafailov, Archit Sharma, Eric Mitchell, Christopher~D Manning, Stefano Ermon, and Chelsea Finn. 2024.
\newblock Direct preference optimization: Your language model is secretly a reward model.
\newblock \emph{Advances in Neural Information Processing Systems}, 36.

\bibitem[{Ratner et~al.(2023)Ratner, Levine, Belinkov, Ram, Magar, Abend, Karpas, Shashua, Leyton-Brown, and Shoham}]{ratner-etal-2023-parallel}
Nir Ratner, Yoav Levine, Yonatan Belinkov, Ori Ram, Inbal Magar, Omri Abend, Ehud Karpas, Amnon Shashua, Kevin Leyton-Brown, and Yoav Shoham. 2023.
\newblock \href {https://doi.org/10.18653/v1/2023.acl-long.352} {Parallel context windows for large language models}.
\newblock In \emph{Proceedings of the 61st Annual Meeting of the Association for Computational Linguistics (Volume 1: Long Papers)}, pages 6383--6402, Toronto, Canada. Association for Computational Linguistics.

\bibitem[{Schulman et~al.(2017)Schulman, Wolski, Dhariwal, Radford, and Klimov}]{schulman2017proximal}
John Schulman, Filip Wolski, Prafulla Dhariwal, Alec Radford, and Oleg Klimov. 2017.
\newblock Proximal policy optimization algorithms.
\newblock \emph{arXiv preprint arXiv:1707.06347}.

\bibitem[{Shi et~al.(2023)Shi, Chen, Misra, Scales, Dohan, Chi, Sch{\"a}rli, and Zhou}]{shi2023large}
Freda Shi, Xinyun Chen, Kanishka Misra, Nathan Scales, David Dohan, Ed~H Chi, Nathanael Sch{\"a}rli, and Denny Zhou. 2023.
\newblock Large language models can be easily distracted by irrelevant context.
\newblock In \emph{International Conference on Machine Learning}, pages 31210--31227. PMLR.

\bibitem[{Shinn et~al.(2024)Shinn, Cassano, Gopinath, Narasimhan, and Yao}]{shinn2024reflexion}
Noah Shinn, Federico Cassano, Ashwin Gopinath, Karthik Narasimhan, and Shunyu Yao. 2024.
\newblock Reflexion: Language agents with verbal reinforcement learning.
\newblock \emph{Advances in Neural Information Processing Systems}, 36.

\bibitem[{Shridhar et~al.(2020)Shridhar, Yuan, C{\^o}t{\'e}, Bisk, Trischler, and Hausknecht}]{shridhar2020alfworld}
Mohit Shridhar, Xingdi Yuan, Marc-Alexandre C{\^o}t{\'e}, Yonatan Bisk, Adam Trischler, and Matthew Hausknecht. 2020.
\newblock Alfworld: Aligning text and embodied environments for interactive learning.
\newblock \emph{arXiv preprint arXiv:2010.03768}.

\bibitem[{Song et~al.(2024{\natexlab{a}})Song, Yu, Li, Yu, Huang, Li, and Wang}]{song2024preference}
Feifan Song, Bowen Yu, Minghao Li, Haiyang Yu, Fei Huang, Yongbin Li, and Houfeng Wang. 2024{\natexlab{a}}.
\newblock Preference ranking optimization for human alignment.
\newblock In \emph{Proceedings of the AAAI Conference on Artificial Intelligence}, volume~38, pages 18990--18998.

\bibitem[{Song et~al.(2024{\natexlab{b}})Song, Yin, Yue, Huang, Li, and Lin}]{song2024trial}
Yifan Song, Da~Yin, Xiang Yue, Jie Huang, Sujian Li, and Bill~Yuchen Lin. 2024{\natexlab{b}}.
\newblock Trial and error: Exploration-based trajectory optimization for llm agents.
\newblock \emph{arXiv preprint arXiv:2403.02502}.

\bibitem[{Talebirad and Nadiri(2023)}]{talebirad2023multi}
Yashar Talebirad and Amirhossein Nadiri. 2023.
\newblock Multi-agent collaboration: Harnessing the power of intelligent llm agents.
\newblock \emph{arXiv preprint arXiv:2306.03314}.

\bibitem[{Wang et~al.(2024{\natexlab{a}})Wang, Wang, Athiwaratkun, Zhang, and Zou}]{wang2024mixture}
Junlin Wang, Jue Wang, Ben Athiwaratkun, Ce~Zhang, and James Zou. 2024{\natexlab{a}}.
\newblock Mixture-of-agents enhances large language model capabilities.
\newblock \emph{arXiv preprint arXiv:2406.04692}.

\bibitem[{Wang et~al.(2024{\natexlab{b}})Wang, Li, Han, Zhang, and Baldwin}]{wang2024learning}
Renxi Wang, Haonan Li, Xudong Han, Yixuan Zhang, and Timothy Baldwin. 2024{\natexlab{b}}.
\newblock Learning from failure: Integrating negative examples when fine-tuning large language models as agents.
\newblock \emph{arXiv preprint arXiv:2402.11651}.

\bibitem[{Wang et~al.(2022)Wang, Wei, Schuurmans, Le, Chi, Narang, Chowdhery, and Zhou}]{wang2022self}
Xuezhi Wang, Jason Wei, Dale Schuurmans, Quoc Le, Ed~Chi, Sharan Narang, Aakanksha Chowdhery, and Denny Zhou. 2022.
\newblock Self-consistency improves chain of thought reasoning in language models.
\newblock \emph{arXiv preprint arXiv:2203.11171}.

\bibitem[{Wei et~al.(2022)Wei, Wang, Schuurmans, Bosma, Xia, Chi, Le, Zhou et~al.}]{wei2022chain}
Jason Wei, Xuezhi Wang, Dale Schuurmans, Maarten Bosma, Fei Xia, Ed~Chi, Quoc~V Le, Denny Zhou, et~al. 2022.
\newblock Chain-of-thought prompting elicits reasoning in large language models.
\newblock \emph{Advances in neural information processing systems}, 35:24824--24837.

\bibitem[{Wei et~al.(2023)Wei, Wang, Liu, Ding, and Zhang}]{wei2023magicoder}
Yuxiang Wei, Zhe Wang, Jiawei Liu, Yifeng Ding, and Lingming Zhang. 2023.
\newblock Magicoder: Source code is all you need.
\newblock \emph{arXiv preprint arXiv:2312.02120}.

\bibitem[{Wu et~al.(2023{\natexlab{a}})Wu, Bansal, Zhang, Wu, Zhang, Zhu, Li, Jiang, Zhang, and Wang}]{wu2023autogen}
Qingyun Wu, Gagan Bansal, Jieyu Zhang, Yiran Wu, Shaokun Zhang, Erkang Zhu, Beibin Li, Li~Jiang, Xiaoyun Zhang, and Chi Wang. 2023{\natexlab{a}}.
\newblock Autogen: Enabling next-gen llm applications via multi-agent conversation framework.
\newblock \emph{arXiv preprint arXiv:2308.08155}.

\bibitem[{Wu et~al.(2024)Wu, Xie, Chen, Zhu, Zhang, and Xiao}]{wu2024easily}
Siye Wu, Jian Xie, Jiangjie Chen, Tinghui Zhu, Kai Zhang, and Yanghua Xiao. 2024.
\newblock How easily do irrelevant inputs skew the responses of large language models?
\newblock \emph{arXiv preprint arXiv:2404.03302}.

\bibitem[{Wu et~al.(2023{\natexlab{b}})Wu, Jia, Zhang, Li, Zhu, Wang, Lee, Peng, Wu, and Wang}]{wu2023empirical}
Yiran Wu, Feiran Jia, Shaokun Zhang, Hangyu Li, Erkang Zhu, Yue Wang, Yin~Tat Lee, Richard Peng, Qingyun Wu, and Chi Wang. 2023{\natexlab{b}}.
\newblock An empirical study on challenging math problem solving with gpt-4.
\newblock In \emph{ArXiv preprint arXiv:2306.01337}.

\bibitem[{Xi et~al.(2024{\natexlab{a}})Xi, Chen, Hong, Jin, Zheng, He, Ding, Liu, Guo, Wang et~al.}]{xi2024training}
Zhiheng Xi, Wenxiang Chen, Boyang Hong, Senjie Jin, Rui Zheng, Wei He, Yiwen Ding, Shichun Liu, Xin Guo, Junzhe Wang, et~al. 2024{\natexlab{a}}.
\newblock Training large language models for reasoning through reverse curriculum reinforcement learning.
\newblock \emph{arXiv preprint arXiv:2402.05808}.

\bibitem[{Xi et~al.(2024{\natexlab{b}})Xi, Ding, Chen, Hong, Guo, Wang, Yang, Liao, Guo, He, Gao, Chen, Zheng, Zou, Gui, Zhang, Qiu, Huang, Wu, and Jiang}]{xi2024agentgym}
Zhiheng Xi, Yiwen Ding, Wenxiang Chen, Boyang Hong, Honglin Guo, Junzhe Wang, Dingwen Yang, Chenyang Liao, Xin Guo, Wei He, Songyang Gao, Lu~Chen, Rui Zheng, Yicheng Zou, Tao Gui, Qi~Zhang, Xipeng Qiu, Xuanjing Huang, Zuxuan Wu, and Yu-Gang Jiang. 2024{\natexlab{b}}.
\newblock \href {http://arxiv.org/abs/2406.04151} {Agentgym: Evolving large language model-based agents across diverse environments}.

\bibitem[{Xie et~al.(2024)Xie, Zhang, Chen, Zhu, Lou, Tian, Xiao, and Su}]{xie2024travelplanner}
Jian Xie, Kai Zhang, Jiangjie Chen, Tinghui Zhu, Renze Lou, Yuandong Tian, Yanghua Xiao, and Yu~Su. 2024.
\newblock Travelplanner: A benchmark for real-world planning with language agents.
\newblock \emph{arXiv preprint arXiv:2402.01622}.

\bibitem[{Xie and Min(2022)}]{xie2022does}
Sang~Michael Xie and Sewon Min. 2022.
\newblock How does in-context learning work? a framework for understanding the differences from traditional supervised learning.

\bibitem[{Yang et~al.(2023)Yang, Liu, Han, Chen, Huang, Fu, and Yu}]{yang2023appagent}
Zhao Yang, Jiaxuan Liu, Yucheng Han, Xin Chen, Zebiao Huang, Bin Fu, and Gang Yu. 2023.
\newblock Appagent: Multimodal agents as smartphone users.
\newblock \emph{arXiv preprint arXiv:2312.13771}.

\bibitem[{Yao et~al.(2022{\natexlab{a}})Yao, Chen, Yang, and Narasimhan}]{yao2022webshop}
Shunyu Yao, Howard Chen, John Yang, and Karthik Narasimhan. 2022{\natexlab{a}}.
\newblock Webshop: Towards scalable real-world web interaction with grounded language agents.
\newblock \emph{Advances in Neural Information Processing Systems}, 35:20744--20757.

\bibitem[{Yao et~al.(2024)Yao, Yu, Zhao, Shafran, Griffiths, Cao, and Narasimhan}]{yao2024tree}
Shunyu Yao, Dian Yu, Jeffrey Zhao, Izhak Shafran, Tom Griffiths, Yuan Cao, and Karthik Narasimhan. 2024.
\newblock Tree of thoughts: Deliberate problem solving with large language models.
\newblock \emph{Advances in Neural Information Processing Systems}, 36.

\bibitem[{Yao et~al.(2022{\natexlab{b}})Yao, Zhao, Yu, Du, Shafran, Narasimhan, and Cao}]{yao2022react}
Shunyu Yao, Jeffrey Zhao, Dian Yu, Nan Du, Izhak Shafran, Karthik Narasimhan, and Yuan Cao. 2022{\natexlab{b}}.
\newblock React: Synergizing reasoning and acting in language models.
\newblock \emph{arXiv preprint arXiv:2210.03629}.

\bibitem[{Zeng et~al.(2023)Zeng, Liu, Lu, Wang, Liu, Dong, and Tang}]{zeng2023agenttuning}
Aohan Zeng, Mingdao Liu, Rui Lu, Bowen Wang, Xiao Liu, Yuxiao Dong, and Jie Tang. 2023.
\newblock Agenttuning: Enabling generalized agent abilities for llms.
\newblock \emph{arXiv preprint arXiv:2310.12823}.

\bibitem[{Zhang et~al.(2024{\natexlab{a}})Zhang, Xin, Li, Zhang, and Liu}]{zhang2024meta}
Cong Zhang, Deik Derrick~Goh Xin, Dexun Li, Hao Zhang, and Yong Liu. 2024{\natexlab{a}}.
\newblock Meta-task planning for language agents.
\newblock \emph{arXiv preprint arXiv:2405.16510}.

\bibitem[{Zhang et~al.(2024{\natexlab{b}})Zhang, Lan, Murthy, Liu, Yao, Tan, Hoang, Yang, Feng, Liu et~al.}]{zhang2024agentohana}
Jianguo Zhang, Tian Lan, Rithesh Murthy, Zhiwei Liu, Weiran Yao, Juntao Tan, Thai Hoang, Liangwei Yang, Yihao Feng, Zuxin Liu, et~al. 2024{\natexlab{b}}.
\newblock Agentohana: Design unified data and training pipeline for effective agent learning.
\newblock \emph{arXiv preprint arXiv:2402.15506}.

\bibitem[{Zhang et~al.(2024{\natexlab{c}})Zhang, Zhang, Liu, Song, Wang, Krishna, and Wu}]{zhang2024training}
Shaokun Zhang, Jieyu Zhang, Jiale Liu, Linxin Song, Chi Wang, Ranjay Krishna, and Qingyun Wu. 2024{\natexlab{c}}.
\newblock Training language model agents without modifying language models.
\newblock \emph{ICML'24}.

\bibitem[{Zhao et~al.(2024)Zhao, Zu, Xu, Lu, He, Ding, Gui, Zhang, and Huang}]{zhao2024longagent}
Jun Zhao, Can Zu, Hao Xu, Yi~Lu, Wei He, Yiwen Ding, Tao Gui, Qi~Zhang, and Xuanjing Huang. 2024.
\newblock Longagent: Scaling language models to 128k context through multi-agent collaboration.
\newblock \emph{arXiv preprint arXiv:2402.11550}.

\bibitem[{Zheng et~al.(2024)Zheng, Gou, Kil, Sun, and Su}]{zheng2024gpt}
Boyuan Zheng, Boyu Gou, Jihyung Kil, Huan Sun, and Yu~Su. 2024.
\newblock Gpt-4v (ision) is a generalist web agent, if grounded.
\newblock \emph{arXiv preprint arXiv:2401.01614}.

\bibitem[{Zhou et~al.(2024{\natexlab{a}})Zhou, Liu, Xu, Iyer, Sun, Mao, Ma, Efrat, Yu, Yu et~al.}]{zhou2024lima}
Chunting Zhou, Pengfei Liu, Puxin Xu, Srinivasan Iyer, Jiao Sun, Yuning Mao, Xuezhe Ma, Avia Efrat, Ping Yu, Lili Yu, et~al. 2024{\natexlab{a}}.
\newblock Lima: Less is more for alignment.
\newblock \emph{Advances in Neural Information Processing Systems}, 36.

\bibitem[{Zhou et~al.(2023)Zhou, Xu, Zhu, Zhou, Lo, Sridhar, Cheng, Bisk, Fried, Alon et~al.}]{zhou2023webarena}
Shuyan Zhou, Frank~F Xu, Hao Zhu, Xuhui Zhou, Robert Lo, Abishek Sridhar, Xianyi Cheng, Yonatan Bisk, Daniel Fried, Uri Alon, et~al. 2023.
\newblock \href {https://webarena.dev} {Webarena: A realistic web environment for building autonomous agents}.
\newblock \emph{arXiv preprint arXiv:2307.13854}.

\bibitem[{Zhou et~al.(2024{\natexlab{b}})Zhou, Zanette, Pan, Levine, and Kumar}]{zhou2024archer}
Yifei Zhou, Andrea Zanette, Jiayi Pan, Sergey Levine, and Aviral Kumar. 2024{\natexlab{b}}.
\newblock Archer: Training language model agents via hierarchical multi-turn rl.
\newblock \emph{arXiv preprint arXiv:2402.19446}.

\bibitem[{Zhu et~al.(2023)Zhu, Wang, Zhou, Wang, Chen, Wang, Yang, Ye, Gong, Zhang et~al.}]{zhu2023promptbench}
Kaijie Zhu, Jindong Wang, Jiaheng Zhou, Zichen Wang, Hao Chen, Yidong Wang, Linyi Yang, Wei Ye, Neil~Zhenqiang Gong, Yue Zhang, et~al. 2023.
\newblock Promptbench: Towards evaluating the robustness of large language models on adversarial prompts.
\newblock \emph{arXiv preprint arXiv:2306.04528}.

\end{thebibliography}

% ==========================================================
% ======================== APPENDIX ========================
% ==========================================================

% \onecolumn
\appendix

\section{Appendix}
\label{sec:appendix}

\renewcommand{\thefigure}{A.\arabic{figure}}
\renewcommand{\thetable}{A.\arabic{table}}
\setcounter{figure}{0}  % reset counter  
\setcounter{table}{0}  % reset counter

\subsection{Dataset}
\label{appendix_dataset}

The TravelPlanner dataset\footnote{TravelPlanner Dataset: \url{https://huggingface.co/datasets/osunlp/TravelPlanner}} consists of three splits of training, validation, and test sets as follows:

\begin{itemize}[noitemsep, leftmargin=*]
    \item The \textbf{Training Set} consists of $45$ triplets of query, reference, and human annotated plan. The annotations are used as demonstrations for in-context learning or supervised fine-tuning in our paper. 
    Please note that these annotated plans are merely a subset of many feasible plans. 
    As expected, the Oracle (i.e., system) returns the feedback for the annotations where no issue is raised (Appx.\ \ref{feedback_example_oracle}).
    \item The \textbf{Validation Set} comes with $180$ pairs of query and reference, with no annotated plans.
    \item The \textbf{Test Set} holds $1,000$ queries together with their references, without any annotated plans.
\end{itemize}

For a given query, agents are expected to formulate a (comprehensive) plan which includes transportation, restaurants, attractions, and accommodation for each day (Appx.\ \ref{ori_example} shows an example).

\subsection{Evaluation metrics}
\label{appendix_eval}

Following TravelPlanner, we use automatic evaluation metrics to assess whether a plan generated by the agent meets the (correct) format condition as well as all the constraints.

\begin{itemize}[noitemsep, leftmargin=*]
    \item \textbf{Delivery Rate} 
    measures whether the agent could successfully generate a plan within a limited number of steps. Falling into any dead loops or invalid plan formats leads to failure. In the sole-planning setting, any failure in drafting a plan negatively impacts the delivery rate.
    \item \textbf{Commonsense Constraint Pass Rate} 
    assesses whether the agent can incorporate commonsense while drafting plans without explicit instructions. For example, the agent has to pick valid entities (incl.\ restaurants, hotels, etc.) from the reference information and not hallucinate.
    \item \textbf{Hard Constraint Pass Rate} 
    measures whether a plan meets all hard constraints mentioned in the query, e.g., budget limit, cuisine preference, or accommodation type. \\
    \underline{N.B.} For Commonsense and Hard Constraint Pass Rates, the evaluation is done in two ways, \textbf{Micro} and \textbf{Macro}, which evaluate the agent's capability of following individual constraints vs.\ all the constraints holistically \cite{xie2024travelplanner}. \\
    \item \textbf{Final Pass Rate} 
    measures whether a plan satisfies all hard and commonsense constraints.
    \item \textbf{Hallucination Rate}
    measures whether a plan contains entities that cannot be found in the reference information.
\end{itemize}

TravelPlanner's Leaderboard\footnote{TravelPlanner Leaderboard: \url{https://huggingface.co/spaces/osunlp/TravelPlannerLeaderboard}} let us evaluate the performance of agents against both validation and test sets \textit{online}. This creates a stage for fair evaluation for all researchers. We use this leaderboard to calculate the figures for the validation and test sets for our experiments. We run each five times, with a different random seed, and report the average scores.

\begin{figure*}[!t]
    \centering
    \resizebox{0.975\linewidth}{!}{
    \includegraphics{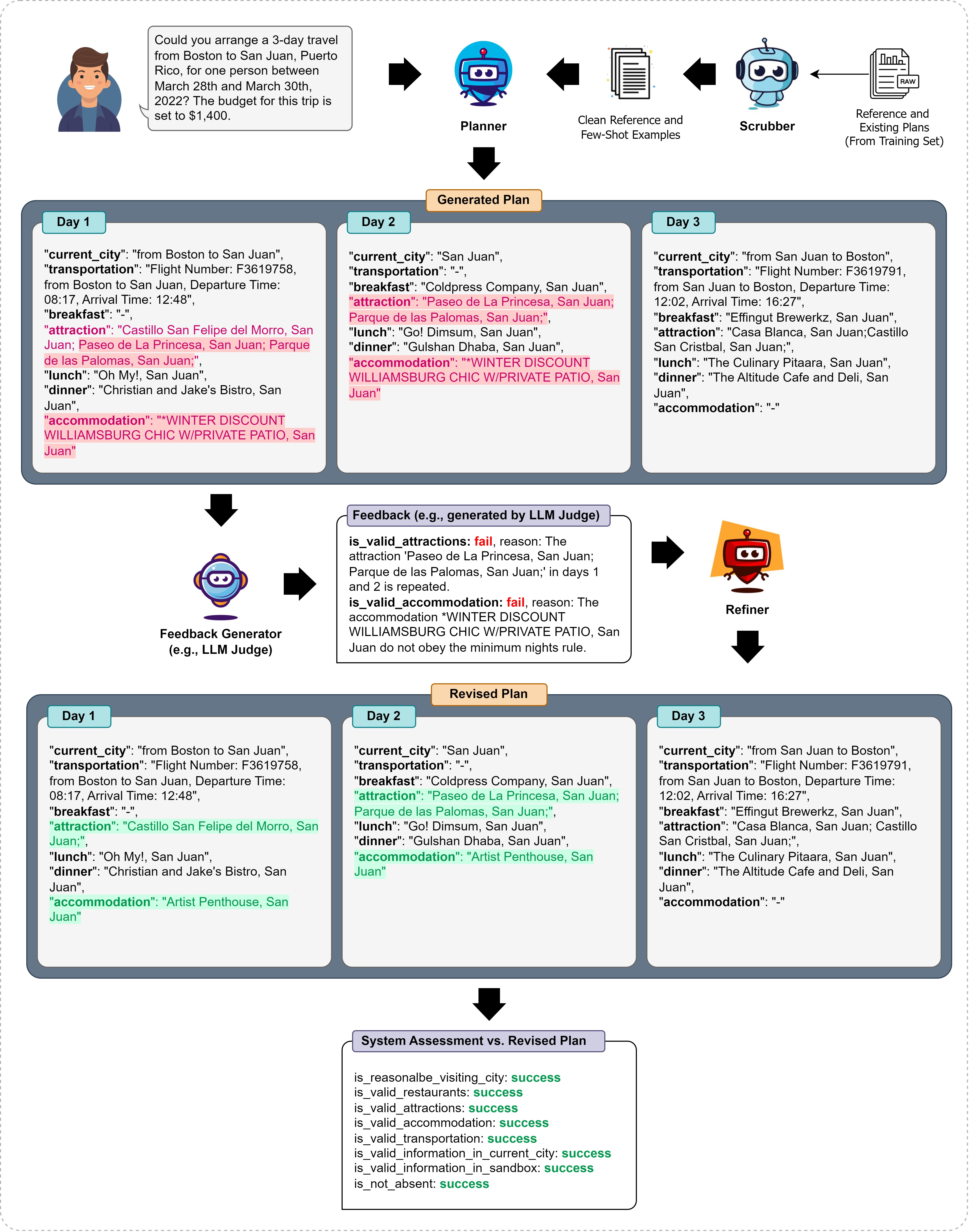}
    }
    \caption{Toy Example: The Planner initially generates a plan w.r.t.\ query and reference, then the feedback generator generates feedback considering the commonsense constraints. Then, the Refiner modifies the plan to meet the requirements for all constraints.}
    \label{appx:fig:toy-example}
\end{figure*}

\subsection{Framework}
\label{appendix_framework}

In Fig.\ \ref{fig:pipeline}, we show that the Planner agent generates a plan for a given query and (cleaned) reference information. In TravelPlanner's \textit{Two-Staging} setting, the reference information is collected by an upstream tool agent which gathers valid information related to transportation, dining, attractions, and accommodation from their corresponding source files. The original benchmark also particularly creates \textit{valid} reference information for the \textit{Sole Planning} setting where the focus is on the Planner agent. Hence, we evaluate our solution only in the \textit{Sole Planning} setting since our focus is on planning.

\subsubsection{The Scrubber Agent}

Since the reference information is massive and lengthy (i.e., $10,000$ tokens on average), we propose the Scrubber, a filtration agent, which infers the hard constraints from the query. There are $5$ hard constraints: {\verb|Room Rule|}, {\verb|Room Type|}, {\verb|Cuisine|}, {\verb|Budget|} and {\verb|Transportation|}. We let the Scrubber to predict the exact constraint value based on the query, for example, one or several cuisine preferences from the set: \{\verb|American|, \verb|Chinese|, \verb|French|, \verb|Indian|, \verb|Italian|, \verb|Mediterranean|, \verb|Mexican|\}. 
Internally within the Scrubber, we inject the whole training set as few-shot examples on top of the test query, to improve the accuracy\footnote{We utilize GPT-4-Turbo and achieve nearly $100\%$ accuracy.}. 
Then, during inference, with the Scrubber agent, each predicted hard constraint is used to remove the rows (from the tables in the reference information) that are not used to produce the final plan. 
For example, if the predicted cuisine preferences are \verb|Italian|, \verb|Mediterranean|, then any restaurants that cannot provide these two cuisine types are removed. So, after removal, the length of reference information becomes shorter. 
Besides, we manually removed several irrelevant columns to the final planning task, such as ratings, phone numbers, and websites, from the tables in the reference. These two efforts massively shorten the long reference information by around $60\%$. It is noteworthy that there are still many choices available to the Planner agent for drafting a correct plan.
For example, if the user's budget for a trip is $\$8,000$, after removing the hotels whose prices are above this limit, there are still other choices left for the Planner agent to reason and draft a plan to meet the budget and other constraints. 
The prompt for the Scrubber agent is found in Appx.\ \ref{prompt_scrubber}.

\subsubsection{The Feedback Generator and Refiner}
\label{refinement_module}

Once the original plan has been drafted, refinement is conducted in an iterative manner.
For this, we follow previous works where two agents are separately created with natural language communication capabilities.

\noindent The \textbf{Feedback Generator} which is responsible for generating nuanced task-dependent feedback that addresses multiple constraints. We tailor a prompt, as shown in Appx.\ \ref{prompt_fb_generator}, to ask LLMs to write feedback with regard to commonsense constraints. In the instructions, we provide a list of constraints with their descriptions. Here two-shots are used to help with feedback generation. The shots are randomly selected from the training set.

\noindent The \textbf{Refiner Agent} refines the generated plan based on the feedback received from the Feedback Generator towards a better version (see the prompt in Appx.\ \ref{prompt_refiner}).

\noindent Fig.\ \ref{appx:fig:toy-example} illustrates the entire refinement phase. The feedback points out that there is a repeated attraction for Days $1$ and $2$, and the accommodation does not satisfy the minimum number of nights requirement.
Then, the Refiner agent refines this draft plan into a new plan where the attraction for the first day is replaced to avoid repetition, and another hotel is chosen which allows a two-night stay. Finally, based on the system assessment, the refined plan meets all commonsense constraints.

% \subsection{SFT and FAFT for Planner}
\subsection{Supervised Fine-Tuning and Feedback-Aware Fine-Tuning} 
\label{appendix_sft_faft}

\subsubsection{Training Example Template for SFT}
The TravelPlanner training set consists of $45$ samples with annotated plans. We use reference information, queries, and annotated plans for general SFT (which is a baseline).

\begin{lstlisting}
reference information box: {ref}
query: {query}
draft travel plan: {plan}
\end{lstlisting}
\label{sft_sample}

\subsubsection{Training Example Template for FAFT}

\begin{lstlisting}
reference information box:{ref}
query:{query}
feedback:{feedback}
draft travel plan:{plan}
\end{lstlisting}
\label{nat_sample}

\subsubsection{Inference Example Template for FAFT}

\begin{lstlisting}
reference information box:{ref}
query:{query}
feedback:{feedback}
is_reasonalbe_visiting_city: success
is_valid_restaurants: success
is_valid_attractions: success
is_valid_accommodation: success
is_valid_transportation: success
is_valid_information_in_current_city: success
is_valid_information_in_sandbox: success
is_not_absent: success
draft travel plan:
\end{lstlisting}
\label{nat_sample_inference}

\subsubsection{Fine-tuning Setup}

In RQ4, for the Planner agent, we fine-tune Llama3-8B for $3$ epochs with a batch size of $4$ for both SFT and FAFT. We use a constant scheduler learning rate of $5\times10^{-5}$ and no warm-up, and we disable packing among training samples to avoid cross-contamination. We train the model in $4$-bit. The maximum sequence length is set to $7000$ to allow the training context to cover all samples. 
For computation and memory efficiency, we also use Low-Rank Adaptation with $r=16$ and $alpha=16$.

\subsection{Prompt Templates for Agents}\label{appendix_prompt_list}
\subsubsection{The Scrubber's Prompt Template}

\begin{lstlisting}
Can you assist in creating a 5-day travel itinerary starting in Sacramento and covering 2 cities in Washington state from March 22nd to March 26th, 2022? The journey will be for a group of three with a budget of $3,600. We require accommodations that provide entire rooms and do not plan to travel by flight. As far as cuisines are concerned, we'd love to experience American, Mediterranean, Italian, and French during our trip.
===> ['American', 'Mediterranean', 'Italian', 'French']

Can you help with generating a 7-day travel plan for a party of 5? We're setting off from Indianapolis and planning to explore 3 cities in Colorado from March 11th to March 17th, 2022. We have a budget of $15,100 for this trip. We'll be bringing our pets, so pet-friendly accommodations are a must. We're also hoping to find places that offer Mexican, Italian, Mediterranean, and Indian cuisines. Entire rooms for accommodations would be ideal.
===> ['Mexican', 'Italian', 'Mediterranean', 'Indian']

Can you assist in creating a travel itinerary for a group of 4, starting in Seattle and visiting 3 unique cities across Texas? This trip will span over 7 days from March 10th through March 16th, 2022. We have a budget of $11,000. Regarding our accommodations, we would like to rent entire rooms, and it's important that our lodgings allow parties. As for transportation, we do not plan to drive ourselves around.
===> []
...{45 shots from trainset}...

I need your help to plan a 5-day vacation for a group of 4 people. We're departing from Honolulu and planning to visit 2 cities in California from March 19th to March 23rd, 2022. The budget for our trip is $11,200. For food preferences, we enjoy Mediterranean and Mexican dishes.
===>{inference for the cuisine preference}
\end{lstlisting}
\label{prompt_scrubber}

\subsubsection{Feedback Generator's Prompt}
\label{prompt_fb_generator}

\begin{lstlisting}
Now You are an advanced reasoning, analyzing and advisory agent who can write feedback and insights for a given draft travel plan, based on the given query and reference information box. 
The feedback you write should check and judge if the given draft travel plan violates one or several following constraints:
* is_reasonalbe_visiting_city: {success or fail}. This refers to  Reasonable City Route: Changes in cities during the trip must be reasonable.
* is_valid_restaurants: {success or fail}. This refers to  Diverse Restaurants: Restaurant choices should not be repeated throughout the trip.
* is_valid_attractions: {success or fail}. This refers to  Diverse Attractions: Attraction choices should not be repeated throughout the trip.
* is_valid_accommodation: {success or fail}. This refers to Minimum Nights Stay: The number of consecutive days spent in a specific accommodation during the trip must meet the corresponding required minimum number of nights' stay.
* is_valid_transportation: {success or fail}. This refers to No conflict Transportation: Transportation choices within the trip must be reasonable. For example, having both "self-driving" and "flight" would be considered a conflict.
* is_valid_information_in_current_city: {success or fail}. This refers to  Within Current City: All scheduled activities for the day must be located within that day's city(s).
* is_valid_information_in_sandbox: {success or fail}. This refers to  Within Sandbox: All information, such as restaurants, attractions, accommodations and transportation, in the plan, must be within the closed sandbox (reference information box); otherwise, it will be considered a hallucination.
* is_not_absent: {success or fail}. This refers to  Complete Information: No key information should be left out of the plan, such as the lack of accommodation during travel.

Here are some examples for your information as demonstrations:

***** Example Starts *****
reference information box:{ref}
query:{query}
draft travel plan:{plan}
feedback:{feedback}
-------------------------------------
reference information box:{ref}
query:{query}
draft travel plan:{plan}
feedback:{feedback}
***** Example Ends *****

Now, You should write the feedback with regard to the aspect of constraints shown above. Follow the formats shown in the examples above.
reference information box:{ref}
query:{query}
draft travel plan:{plan}
feedback:
\end{lstlisting}

\subsubsection{The Refiner's Prompt Template}

\begin{lstlisting}
You are a proficient planner. Based on the provided information and query, please give me a detailed plan, including specifics such as flight numbers (e.g., F0123456), restaurant names, and accommodation names. Note that all the information in your plan should be derived from the provided data. You must adhere to the format given in the example. Additionally, all details should align with commonsense. The symbol '-' indicates that information is unnecessary. For example, in the provided sample, you do not need to plan after returning to the departure city. When you travel to two cities in one day, you should note it in the 'Current City' section as in the example (i.e., from A to B).

***** Example *****
Query: Could you create a travel plan for 7 people from Ithaca to Charlotte spanning 3 days, from March 8th to March 14th, 2022, with a budget of $30,200?
Travel Plan:
Day 1:
Current City: from Ithaca to Charlotte
Transportation: Flight Number: F3633413, from Ithaca to Charlotte, Departure Time: 05:38, Arrival Time: 07:46
Breakfast: Nagaland's Kitchen, Charlotte
Attraction: The Charlotte Museum of History, Charlotte
Lunch: Cafe Maple Street, Charlotte
Dinner: Bombay Vada Pav, Charlotte
Accommodation: Affordable Spacious Refurbished Room in Bushwick!, Charlotte

Day 2:
Current City: Charlotte
Transportation: -
Breakfast: Olive Tree Cafe, Charlotte
Attraction: The Mint Museum, Charlotte; Romare Bearden Park, Charlotte.
Lunch: Birbal Ji Dhaba, Charlotte
Dinner: Pind Balluchi, Charlotte
Accommodation: Affordable Spacious Refurbished Room in Bushwick!, Charlotte

Day 3:
Current City: from Charlotte to Ithaca
Transportation: Flight Number: F3786167, from Charlotte to Ithaca, Departure Time: 21:42, Arrival Time: 23:26
Breakfast: Subway, Charlotte
Attraction: Books Monument, Charlotte.
Lunch: Olive Tree Cafe, Charlotte
Dinner: Kylin Skybar, Charlotte
Accommodation: -

***** Example Ends *****

Given information: {reference information box}
Query: {query}
Travel Plan: {original draft travel plan}

Now You are an advanced reasoning and self-corrective agent that can improve based on self refection and the feedback. 
Based on given query, reference information box, and proposed Travel Plan, above, you are now given the feedback which includes the reason why it fails.
Try to write a new plan in which the errors are fixed.
Keep in mind that you only make changes or replace the item which causes the issue.
If it appears at multiple places, correct them all at once.
Try to avoid making unnecessary changes on the previous proposed plan. 
Always make sure that your generation, such as the names of resuturants, attractions, accommodations, transportations, can be found in the given reference information box above.
For attraction, breakfast, dinner and lunch, do not give repetition within each day and among the days in the plan, i.e. each of them should NOT appear more than once in the whole travel plan.
Feel free to ignore irrelevant information in reference information box.

* If the feedback is about repeated restaurant, for example, "The restaurant in day 4 dinner is repeated.", then you need to take another restuartant from reference infobox, which is different from the previous one and all other chosen ones in the plan;
* If the feedback is about "The breakfast/lunch/dinner/attraction/accommodation in day X is invalid in the sandbox", for example, "The lunch in day 3 is invalid in the sandbox.", this means that the choice cannot be found in reference infobox. Then you should take another one which is definitely within the inference infobox.
* If the feedback is about "The accommodation X do not obey the minumum nights rule",   this means that the total days/nights spent in the accommodation place chosen in the plan, does not obey the minumum nights rule.

    For example, if the days spent in that accommodation in the plan are 2 days, but the 'minumum nights' of that accommodation is greater than 2, then the plan violates the rule. 
    Therefore, you should review and examine the number of 'minumum nights' of each accommodation in the reference information box and make sure the days spent in that accommodation is equal or greater than that number.
    
* If the feedback is about "No accommodation/transportation/attaction/meal in day X is not allowed", this means that on that day, you should arrange the corresponding activity rather than leave it blank(denoted as '-').
* If the feedback is about "The transportation is conflicting.", this means that you cannot select neither the combination of Taxi and Self-driving nor the combination of Flight and Self-driving, at the same time, in terms of transportation.

Feedback: {feedback}

Write a new plan:
\end{lstlisting}
\label{prompt_refiner}

\subsection{Case Presentation}\label{appendix_case_presentaion}

% \subsubsection{An example of query with its corresponding travel plan}
% \subsubsection{Query Example with its Corresponding Travel Plan}
\subsubsection{Query Example with its Travel Plan}

\begin{lstlisting}
QUERY:
Can you create a travel plan for a group of 4 departing from Seattle 2 and heading to San Francisco for 3 days, from March 6 th to March 8th,2022? Our budget is $2,900. We are bringing pets, so accommodations need to be pet-friendly. We are interested in trying Mexican, French, American, and Mediterranean cuisines during our visit. We would also prefer to avoid flying for transportation.

TRAVEL PLAN:
Day 1: 
Current City: from Seattle to San Francisco
Transportation: Self-Driving from Seattle to San Francisco, Duration: 12 hours 28 mins, Cost: $65
Breakfast: - 
Attraction: - 
Lunch: -
Dinner: Anupam Eating Point, San Francisco
Accommodation: Room in Down town Brooklyn Parkslop, San Francisco

Day 2: 
Current City: San Francisco 
Transportation: - 
Breakfast: Coffee & Chai Co., San Francisco 
Attraction: Golden Gate Bridge, San Francisco; Golden Gate Park, San Francisco
Lunch: Bonne Bouche, San Francisco
Dinner: Empress, San Francisco
Accommodation: Room in Down town Brooklyn Parkslop, San Francisco

Day 3: 
Current City: from San Francisco to Seattle 
Transportation: Self-Driving from San Francisco to Seattle, Duration :12 hours 25 mins, Cost: $65
Breakfast: Gupta's Rasoi, San Francisco
Attraction: PIER 39, San Francisco
Lunch: Shammi Bhai Lassi Wala, San Francisco
Dinner: -
Accommodation: -
\end{lstlisting}
\label{ori_example}

% \subsubsection{An example of the feedback from LLMs}
\subsubsection{Feedback Examples Generated by LLMs}

% The feedback generated by LLMs are in alignment with the format of the system feedback.
The feedback generated by LLMs is in the same format of the system feedback.

\begin{lstlisting}
is_reasonalbe_visiting_city: fail, reason:The trip should be a closed circle.
is_valid_restaurants: success
is_valid_attractions: success
is_valid_accommodation: fail, reason:The accommodation Harlem cozy nights, Denver(Colorado) do not obey the minumum nights rule.
is_valid_transportation: fail, reason:The transportation is conflicting.
is_valid_information_in_current_city: success
is_valid_information_in_sandbox: fail, reason:The accommodation in day 3 is invalid in the sandbox.
is_not_absent: success
\end{lstlisting}
\label{appx:feedback-example}

\noindent Recall that, there are $45$ annotated plans in the training set. For each plan, without any exception, the generated feedback from the Oracle system is \verb|all-success|:

\begin{lstlisting}
is_reasonalbe_visiting_city: success
is_valid_restaurants: success
is_valid_attractions: success
is_valid_accommodation: success
is_valid_transportation: success
is_valid_information_in_current_city: success
is_valid_information_in_sandbox: success
is_not_absent: success
\end{lstlisting}
\label{feedback_example_oracle}

% \newpage

\begin{table*}[ht]
    \centering
    \resizebox{0.9\linewidth}{!}{
        \begin{tabular}{llcccc}
        \toprule
        \textbf{Benchmark} & \textbf{Task} & \begin{tabular}[c]{@{}c@{}}\textbf{Feedback} \\ \textbf{Provided?}\end{tabular} & \begin{tabular}[c]{@{}c@{}}\textbf{Trajectory} \\ \textbf{Released?}\end{tabular} & \textbf{Baseline} & \begin{tabular}[c]{@{}c@{}}\textbf{Realistic} \\ \textbf{Interface?}\end{tabular} \\ \midrule
        WebShop \cite{yao2022webshop} & \colorbox{my_red}{Web} & No & Expert & Rule, IL, RL, IL+RL & Yes \\ \midrule
        WebArena \cite{zhou2023webarena} & \colorbox{my_red}{Web} & No & Expert, Agent & Direct, CoT & Yes \\ \midrule
        AgentBench \cite{liu2023agentbench} & \begin{tabular}[c]{@{}l@{}}\colorbox{my_red}{Web}, \colorbox{my_green}{Code}, \\ \colorbox{my_yellow}{Game}, \colorbox{my_purple}{Embodiment} \end{tabular} & No & Not Found & CoT & Yes \\ \midrule
        TravelPlanner \cite{xie2024travelplanner} & \colorbox{my_blue}{Tool}, \colorbox{my_orange}{Planning} & Yes & Expert & \begin{tabular}[c]{@{}c@{}}Direct, CoT,\\ ReAct, Reflexion\end{tabular} & No \\ \midrule
        AgentBoard \cite{ma2024agentboard} & \begin{tabular}[c]{@{}l@{}}\colorbox{my_red}{Web}, \colorbox{my_yellow}{Game}, \colorbox{my_blue}{Tool},\\ \colorbox{my_purple}{Embodiment} \end{tabular} & Partially & Not Found & Direct & Partially \\ \midrule
        AgentGym \cite{xi2024agentgym} & \begin{tabular}[c]{@{}l@{}}\colorbox{my_red}{Web}, \colorbox{my_green}{Code}, \colorbox{my_yellow}{Game}, \\  \colorbox{my_blue}{Tool}, \colorbox{my_purple}{Embodiment} \end{tabular} & Partially & Expert, Agent & \begin{tabular}[c]{@{}c@{}}BC (SFT),\\ ReAct, AGENTEVOL\end{tabular} & Yes \\
        \bottomrule
        \end{tabular}
    }
    \caption{Popular Benchmarks for LLM Agents}
    \label{appendix_benchmar_suites}
\end{table*}

\subsection{Related Works}
\label{sec:related-works}

\subsubsection{Benchmarks for LLM-based Generalist Agents}

It has been anticipated that generalist agents can handle diverse tasks and evolve across different (cyber) environments at the human level which is a long-term goal in the AGI community. 
LLMs can be used as experts, which mimic humans, that have a strong generalization capability that not only suits conventional NLP but also agentic tasks.
Recently, plenty of benchmarks have been proposed to evaluate the agents across various tasks and environments comprehensively and fairly. 
We provide an overview of popular benchmarks in the community in Table \ref{appendix_benchmar_suites}.
Some benchmarks such as  ALFWorld \cite{shridhar2020alfworld} and Mind2Web \cite{deng2023mind2web}, which are already included in larger benchmarks, are not listed in the table. 
Although the recent progress in multi-modal LLMs has spurred research into multi-modal LLM agents \cite{yang2023appagent,zheng2024gpt}, we only list benchmarks that focus exclusively on text-based environments which assess LLM agents' abilities via textual reasoning and taking actions in-depth.

The listed benchmarks support agents powered by both API-based proprietary and open-weight LLMs with convenient drop-in replacement interfaces. It is also free to add few-shots or use 
other prompting strategies to generate actions.

\subsubsection{Long Contexts Challenge for LLMs}

Besides the fact that more and more LLMs offer long-context capabilities \cite{fei2023extending,ratner-etal-2023-parallel, liu2023tcra, zhao2024longagent, qian2024long}, recent studies question LLMs' ability to find needles in a haystack because they face challenges in discriminating highly semantically related information, and can be easily distracted by irrelevant and misleading contents in long contexts \cite{wu2024easily, zhu2023promptbench, chang2024survey, shi2023large, LLMTest_NeedleInAHaystack}.
The TravelPlanner \cite{xie2024travelplanner} is a benchmark to provide insightful answers to this problem, wherein lengthy context information, noise, and relevant snippets are deeply intertwined.

\subsubsection{Multi-Agent Collaboration}

Recent studies have borrowed the multiple-agent methodology for collaboration on cyber tasks, gaming, coding, math reasoning, conversation responding, and question answering \cite{guo2024large,wu2023autogen,wu2023empirical,zhang2024training,li2024camel,liu2024agentlite,talebirad2023multi,zhang2024meta,wang2024mixture}. 
Under the hood, these works assign role-specific prompts to the LLM to build multiple agents for synergy and collaboration. 
The self-refinement works can be classified into this realm, where the advisor and refiner agents can troubleshoot and modify the response in a few rounds \cite{madaan2024self,paul2023refiner,kim2024language,pan2024autonomous,chen2023improving}. However, few works study the reliability and robustness of multi-agent collaboration in more complex and practical tasks.
Compared to the previous testbeds where 
% errors in the generations 
generation errors
are easily noticeable and unambiguous, it is questionable whether refinement can work on TravelPlanner, where the glitches are hard to find due to implicit commonsense constraints. 
Multi-agent collaboration also places higher demands on the capabilities of individual agents since a failure at any stage from any agent can lead to a collapse, such as a dead loop or deviation from the goal.

\subsubsection{Reinforcement Learning via Feedback}

On top of works that only use successful trajectories for behavioural cloning \cite{zeng2023agenttuning,chen2023fireact,zhang2024agentohana,chen2024agent}, 
another line of work trains LLM-based agents based on environmental feedback, referred to as
interactive learning methods \cite{song2024trial, zhou2024archer, christianos2023pangu, xi2024training}. 
Specifically, they train the agents via reinforcement learning. However, poor transferability among scenarios, reward inconsistency, off-policy shift, step-level reward sparsity, and training stability and expenses are the main roots of performance bottlenecks.
Possible alternative approaches such as Negative Aware Training (NAT) \cite{wang2024learning, Li2023TurningDI} can be a more robust solution. Our FAFT approach is motivated by NAT, and it can be seamlessly migrated to other agentic tasks. 
    
\end{document}